# On the Intertranslatability of Argumentation Semantics

**Wolfgang Dvořák**                           DVORAK@DBAI.TUWIEN.AC.AT
**Stefan Woltran**                            WOLTRAN@DBAI.TUWIEN.AC.AT
*Technische Universität Wien,*
*Institute of Information Systems 184/2*
*Favoritenstrasse 9-11, 1040 Vienna, Austria*

## Abstract

Translations between different nonmonotonic formalisms always have been an important topic in the field, in particular to understand the knowledge-representation capabilities those formalisms offer. We provide such an investigation in terms of different semantics proposed for abstract argumentation frameworks, a nonmonotonic yet simple formalism which received increasing interest within the last decade. Although the properties of these different semantics are nowadays well understood, there are no explicit results about intertranslatability. We provide such translations wrt. different properties and also give a few novel complexity results which underlie some negative results.

## 1. Introduction

Studies on the intertranslatability of different approaches to nonmonotonic reasoning have always been considered as an important contribution to the field in order to understand the expressibility and representation capacity of the various formalisms. By intertranslatability we understand a function $Tr$ which maps theories from one formalism into another such that intended models of a theory $\Delta$ from the source formalism are in a certain relation to the intended models of $Tr(\Delta)$. Several desired properties for such translation functions have been identified, including to be polynomial ($Tr(\Delta)$ can be computed in polynomial time wrt. the size of $\Delta$) or to be modular (roughly speaking, that allows to transform parts of the theory independently of each other). In particular, the relationship between (variants of) default logic (Reiter, 1980) and nonmonotonic modal logics, e.g. autoepistemic logic (Moore, 1985), has always received a lot of attention, (see, e.g., Denecker, Marek, & Truszczyński, 2003; Konolige, 1988; Marek & Truszczyński, 1993). Perhaps most notably, Gottlob (1995) showed that a modular translation from default logic to autoepistemic logic is impossible. Other important contributions in this direction include translations between default logic and circumscription (Imielinski, 1987), modal nonmonotonic logics and logic programs (see, e.g., de Bruijn, Eiter, & Tompits, 2008 for an overview and recent applications) and the work by Janhunen (1999). Let us also refer here to recent work by Pearce and Uridia (2011), who show that translations of the aforementioned kind have already been known in the context of non-classical logics and related results date back to the work of Gödel.

In this work, we study translation functions within a particular formalism of nonmonotonic reasoning but wrt. to different semantics proposed for this formalism. In the area of default logic, similar research was undertaken, for instance by Liberatore (2007) or Delgrande and Schaub (2005). Likewise, for work concerning the relationship between different logic programming semantics we refer to the work of Janhunen, Niemelä, Seipel, Simons, and You (2006) and the references therein. The formalism we focus on in this paper are Dung's argumentation frameworks (Dung, 1995) which received increasing interest within the last decade. In a nutshell, such argumentation frameworks (AFs, for short) represent abstract statements[1] together with a relation denoting attacks between them. Different semantics provide different ways to solve the inherent conflicts between statements by selecting acceptable subsets — usually called extensions — of them. Several such semantics have already been proposed by Dung in his seminal paper (Dung, 1995), but also alternative approaches play a major role nowadays (see, e.g., Baroni, Dunne, & Giacomin, 2011; Baroni, Giacomin, & Guida, 2005; Caminada, 2006; Dung, Mancarella, & Toni, 2007; Verheij, 1996). Compared to other nonmonotonic formalisms (which are built on top of classical logical syntax), argumentation frameworks are a much simpler formalism (in the end, they are just directed graphs). However, this simplicity made them an attractive modeling tool in several diverse areas, like formalizations of legal reasoning (Bench-Capon & Dunne, 2005) or multi-agent negotiation (Amgoud, Dimopoulos, & Moraitis, 2007).

In the field of argumentation, intertranslatability has mainly been studied in connection with generalizations of Dung's argumentation frameworks. By generalization we mean here the augmentation of simple frameworks by further concepts as priorities or additional relations between arguments. In this context, translations were used to show that proposed semantics for such generalizations are in a close relation with the corresponding semantics of standard AFs. In other words, given such a generalized AF one is interested in translating them to standard AFs while preserving semantics. Such translations have been discussed, for instance, in terms of bipolar AFs (Cayrol & Lagasquie-Schiex, 2009), value-based AFs (Bench-Capon & Atkinson, 2009), AFs with recursive attacks (Baroni, Cerutti, Giacomin, & Guida, 2011), or abstract dialectical frameworks (Brewka, Dunne, & Woltran, 2011). A recent exception where intertranslatability within Dung AFs is discussed, is the work by Baumann and Brewka (2010), who consider to enforce a desired extension in Dung AFs by adding new arguments and switching semantics. From a slightly different perspective, also the work by Gabbay (2009) is related, since it investigates the substitution of an argumentation framework as a node in another framework.

We focus here exclusively on standard argumentation frameworks and have the following main objective: Given an AF $F$ and argumentation semantics $\sigma$ and $\sigma'$, find a function $Tr$ such that the $\sigma$-extensions of $F$ are in certain correspondence to the $\sigma'$-extensions of $Tr(F)$. We believe that such results are important from different points of view.

Firstly, consider there is an advanced argumentation engine for a semantics $\sigma'$, but one wants to evaluate on a different semantics $\sigma$. Then, it might be a good plan to transform $F$ in such a way into an AF $F'$ such that evaluating $F'$ wrt. semantics $\sigma'$ allows for an easy reconstruction of the $\sigma$-extensions of $F$. If the required transformations are efficiently computable, this leads to a potentially more successful approach than implementing a distinguished algorithm for the $\sigma$-semantics from scratch. Figure 1 illustrates this idea. The concept of a filter is required in case $Tr(F)$ introduces further arguments (which thus might appear in the $\sigma'$-extensions of $Tr(F)$) in the course of the translation making a filter-ing of these new arguments necessary to obtain the desired original extensions. However, we will also consider translations for which such a back-translation is not necessary.

---

1. In general, arguments are not considered as simple statements but contain a number of reasons that lead to a conclusion (see, e.g., Besnard & Hunter, 2001; Caminada & Amgoud, 2007). However, for the purpose of this work, we will treat arguments as atomic entities thus abstracting from their internal structure.

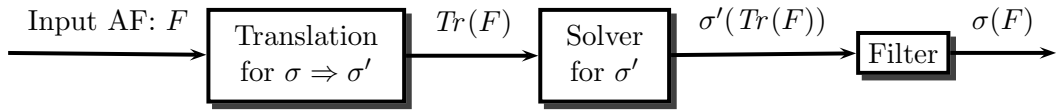

Figure 1: Solver for semantics $\sigma$.

A second motivation of our work is concerned with meta(level) argumentation (see, e.g., Modgil & Bench-Capon, 2011; Villata, 2010) which can be explained as follows: "A meta-level Dung argumentation framework is itself instantiated by arguments that make statements about arguments, their interactions, and their evaluation in an object-level argumentation framework" (Modgil & Bench-Capon, 2011). The translations we shall present here exactly fit into this picture in the sense that we have to model certain features of a semantics $\sigma$ within another semantics $\sigma'$, when giving a translation from $\sigma$ to $\sigma'$. To have a more concrete example, let $\sigma'$ be the complete semantics and $\sigma$ denote stable semantics (we will provide the formal details about the different semantics in Section 2; for the sake of this illustration the details are not required). Then, the transformation has to capture the concept of admissibility (informally speaking, a set of arguments has to defend itself) which is implicitly present in complete semantics by a suitable introduction of new arguments, such that stable semantics can perform such a type of reasoning. In other words, translatability results between different semantics of AFs yield an understanding of how certain properties, which are specified implicitly within one semantics, can be made (syntactically) explicit within an AF in order to make these properties amenable to another semantics.

As a third important application of our work, we consider situations where different semantics of argumentation have to be dealt with simultaneously. This might be the case if different agents share their views about a certain situation (modeled as AFs) but these agents use different semantics to reason over their frameworks. As well, the problem of combining frameworks which have been constructed under the assumption that they will be evaluated under different semantics falls into a possible application area of our work.

Finally, we emphasize that understanding which translations can be efficiently performed wrt. different semantics complements the picture about expressibility of argumentation semantics. For instance, if there exists an efficient translation from semantics $\sigma$ to semantics $\sigma'$, but there is no such translation in the other direction, $\sigma$ could be understood as more expressible than $\sigma'$, although complexity analysis for typical decision problems associated to AFs does not show any difference between $\sigma$ and $\sigma'$. As an example consider semi-stable and stage semantics. For both semantics the credulous acceptance problem is $\Sigma_2^P$-complete and the skeptical acceptance problem is $\Pi_2^P$-complete (Dvořák & Woltran, 2010). But when considering what we call efficient exact translations one can map stage semantics to semi-stable semantics but not vice versa. Thus semi-stable semantics are more expressible than stage semantics wrt. efficient exact translations. However we will argue that the notion of exact translations may be too restrictive for comparing the expressibility of argumentation semantics.

Beside these aspects of motivation, we would like to mention that positive results on intertranslations indicate a certain form of independence of semantics in argumentation. This is of particular importance, having in mind that the argumentation community nowadays is overwhelmed with different proposals of semantics. Thus understanding the basic principles underlying different semantics is crucial, and we believe the results provided in this paper contribute to this question.

The organization of the remainder of the paper and its main contributions are as follows:

- In Section 2, we introduce argumentation frameworks and the different semantics we deal with in this paper. We also review known complexity results which we complement in the sense that we show some of the known tractable problems to be $P$-hard; a fact we will use for some impossibility results in Section 5.

- Section 3 defines properties for translations basically along the lines of Janhunen (1999). In particular, we consider here as desired properties efficiency (the translation can be computed in logarithmic space wrt. the given AF), modularity (the translation can be done independently for certain parts of the framework) and faithfulness (there should be a clear correspondence between the extensions of the translated AF and the original AF). However, we also consider some additional features which are needed to deal with some of the argumentation semantics (for instance, the admissible semantics always yields the empty set as one solution; thus filtering such an entire solution is necessary).

- Section 4 contains our main results, in particular we provide translations between Dung's original semantics (admissible, preferred, stable, complete, grounded), stage semantics (Verheij, 1996) and semi-stable semantics (Caminada, 2006). We analyze these translations wrt. the properties mentioned above using as minimal desiderata efficiency and (a particular form of) faithfulness.

- As already mentioned, Section 5 then provides negative results, i.e. we show that certain translations between semantics are not possible. Some of these impossibility results make use of typical complexity-theoretic assumptions; others are genuine due to the different properties of the compared semantics.

- Finally, in Section 6 we conclude the paper with a summary and discussion of the presented results. As well, an outlook to potential future work is given there.

## 2. Argumentation Frameworks

In this section we introduce (abstract) argumentation frameworks (Dung, 1995) and recall the semantics we study in this paper (see also Baroni & Giacomin, 2009, for an overview). Moreover, we highlight and complement complexity results for typical decision problems associated to such frameworks.

**Definition 1.** *An argumentation framework (AF) is a pair $F = (A, R)$ where $A$ is a non-empty set of arguments* [2] *and $R \subseteq A \times A$ is the attack relation. For a given AF $F = (A, R)$* we use $A_F$ to denote the set $A$ of its arguments and $R_F$ to denote its attack relation $R$. The pair $(a, b) \in R$ means that $a$ attacks $b$.

---

2. For technical reasons we only consider AFs with $A \neq \emptyset$.

We sometimes use the notation $a \rightarrowtail^R b$ instead of $(a, b) \in R$. For $S \subseteq A$ and $a \in A$, we also write $S \rightarrowtail^R a$ (resp. $a \rightarrowtail^R S$) in case there exists an argument $b \in S$, such that $b \rightarrowtail^R a$ (resp. $a \rightarrowtail^R b$). In case no ambiguity arises, we use $\rightarrowtail$ instead of $\rightarrowtail^R$.

An AF can naturally be represented as a directed graph. Semantics for argumentation frameworks are given via a function $\sigma$ which assigns to each AF $F = (A, R)$ a set $\sigma(F) \subseteq 2^A$ of extensions. We shall consider here for $\sigma$ the functions $stb$, $adm$, $prf$, $com$, $grd$, $stg$, and $sem$ which stand for stable, admissible, preferred, complete, grounded, stage, and respectively, semi-stable semantics. Before giving the actual definitions for these semantics, we require a few more formal concepts.

**Definition 2.** *Given an AF $F = (A, R)$, an argument $a \in A$ is* defended *(in $F$) by a set $S \subseteq A$ if for each $b \in A$, such that $b \rightarrowtail a$, also $S \rightarrowtail b$ holds. Moreover, for a set $S \subseteq A$, we define the* range *of $S$, denoted as $S_R^+$, as the set $S \cup \{b \mid S \rightarrowtail b\}$.*

We continue with the definitions of the considered semantics. Observe that their common feature is the concept of conflict-freeness, i.e. arguments in an extension are not allowed to attack each other.

**Definition 3.** *Let $F = (A, R)$ be an AF. A set $S \subseteq A$ is* conflict-free *(in $F$), if there are no $a, b \in S$, such that $(a, b) \in R$. For such a conflict-free set $S$, it holds that*

- $S \in stb(F)$, *if for each $a \in A \setminus S$, $S \rightarrowtail a$, i.e. $S_R^+ = A$;*

- $S \in adm(F)$, *if each $a \in S$ is defended by $S$;*

- $S \in prf(F)$, *if $S \in adm(F)$ and there is no $T \in adm(F)$ with $T \supset S$;*

- $S \in com(F)$, *if $S \in adm(F)$ and for each $a \in A$ that is defended by $S$, $a \in S$;*

- $S \in grd(F)$, *if $S \in com(F)$ and there is no $T \in com(F)$ with $T \subset S$;*

- $S \in stg(F)$, *if there is no conflict-free set $T$ in $F$, such that $T_R^+ \supset S_R^+$;*

- $S \in sem(F)$, *if $S \in adm(F)$ and there is no $T \in adm(F)$ with $T_R^+ \supset S_R^+$.*

*For all semantics $\sigma$, the sets defined above are the only ones in $\sigma(F)$.*

We recall that for each AF $F$,

$$stb(F) \subseteq sem(F) \subseteq prf(F) \subseteq com(F) \subseteq adm(F)$$

holds, and that for each of the considered semantics $\sigma$ except stable semantics, $\sigma(F) \neq \emptyset$ holds. The grounded semantics always yields exactly one extension. Moreover if an AF has at least one stable extension then its stable, semi-stable, and stage extensions coincide.

*Example* 1. Consider the AF $F = (A, R)$, with $A = \{a, b, c, d, e\}$ and $R = \{(a, b), (c, b), (c, d), (d, c), (d, e), (e, e)\}$. The graph representation of $F$ is given as follows.

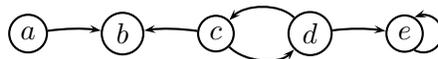

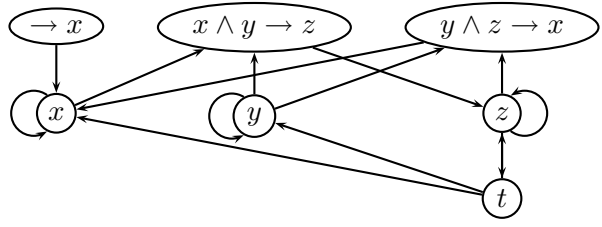

Figure 2: Argumentation framework $F_{T,z}$ for $T = \{\to x, x \wedge y \to z, y \wedge z \to x\}$.

We have $stb(F) = stg(F) = sem(F) = \{\{a, d\}\}$. Further we have as admissible sets of $F$ the collection $\{\}$, $\{a\}$, $\{c\}$, $\{d\}$, $\{a, c\}$, $\{a, d\}$, thus $prf(F) = \{\{a, c\}, \{a, d\}\}$. Finally the complete extensions of $F$ are $\{a\}$, $\{a, c\}$ and $\{a, d\}$, with $\{a\}$ being the grounded extension of $F$. ◇

We now turn to the complexity of reasoning in AFs. To this end, we define the following decision problems for the semantics $\sigma$ introduced in Definition 3.

- *Credulous Acceptance* $\mathsf{Cred}_\sigma$: Given AF $F = (A, R)$ and an argument $a \in A$. Is $a$ contained in some $S \in \sigma(F)$?

- *Skeptical Acceptance* $\mathsf{Skept}_\sigma$: Given AF $F = (A, R)$ and an argument $a \in A$. Is $a$ contained in each $S \in \sigma(F)$?

- *Verification of an extension* $\mathsf{Ver}_\sigma$: Given AF $F = (A, R)$ and a set of arguments $S \subseteq A$. Is $S \in \sigma(F)$?

- *Existence of an extension* $\mathsf{Exists}_\sigma$: Given AF $F = (A, R)$. Is $\sigma(F) \neq \emptyset$?

- *Existence of a nonempty extension* $\mathsf{Exists}_\sigma^{\neg\emptyset}$: Given AF $F = (A, R)$. Does there exist a set $S \neq \emptyset$ such that $S \in \sigma(F)$?

Before giving an overview about known results, we provide a few lower bounds which, to the best of our knowledge, have not been established yet.

**Proposition 1.** *The problems* $\mathsf{Cred}_{grd} = \mathsf{Skept}_{grd} = \mathsf{Skept}_{com}$ *as well as* $\mathsf{Ver}_{grd}$ *are* P*-hard (under* L*-reductions, i.e. reductions using logarithmic space).*

*Proof.* We use a reduction from the P-hard problem to decide, given a propositional definite Horn theory $T$ and an atom $x$, whether $x$ is true in the minimal model of $T$.

Let, for a definite Horn theory $T = \{r_l : b_{l,1} \wedge \cdots \wedge b_{l,i_l} \to h_l \mid 1 \leq l \leq n\}$ over atoms $X$ and an atom $z \in X$, $F_{T,z} = (A, R)$ be an AF defined as follows:

$$A = T \cup X \cup \{t\}$$
$$R = \{(x, x), (t, x) \mid x \in X\} \cup \{(z, t)\} \cup$$
$$\{(r_l, h_l), (b_{l,j}, r_l) \mid r_l \in T, 1 \leq j \leq i_l\}$$

where $t$ is a fresh argument. See Figure 2 for an example. Clearly the AF $F_{T,z}$ can be constructed using only logarithmic space in the size of $T$.

In the following we show that $z$ is in the minimal model of $T$ iff $t$ is in the grounded extension of $F_{T,z}$ iff $grd(F_{T,z}) = \{T \cup \{t\}\}$.

First we attend that $t$ is in the grounded extension $E$ of $F_{T,z}$ iff $E = \{T \cup \{t\}\}$. Obviously the if-direction holds. Thus let us assume $t \in E$, then each $x \in X$ is attacked by $E$ and thus each $r \in T$ is defended by $E$. Hence $E = \{T \cup \{t\}\}$.

It remains to show that $z$ is in the minimal model of $T$ iff $t$ is in the grounded extension $E$ of $F_{T,z}$. We recall the definition of the characteristic function $\mathcal{F}_F$ of an AF $F$, defined as $\mathcal{F}_F(S) = \{x \in A_F \mid x \text{ is defended by } S\}$, and that the grounded extension of $F$ is the least fix-point of $\mathcal{F}_F$. To show the only-if part, let us assume that $z$ is in the minimal model of $T$. Thus there exists a finite sequence of rules $(r_{l_i})_{1 \leq i \leq k}$, such that (i) for each rule $r_{l_i}$ and each atom $b_{l_i,s}$ there exists a rule $r_{l_j}, j < i$ with $h_{l_j} = b_{l_i,s}$ and (ii) $h_{l_k} = z$. Clearly $r_{l_1}$ has empty body and thus the corresponding argument has no attackers in $F_{T,z}$, i.e. $r_{l_1} \in E$. We now claim that for each $i, 1 \leq i \leq k, r_{l_i} \in E$ holds as well and prove this by induction. To this end, we assume the claim holds for all $m < i$, i.e. $r_{l_m} \in E$, and thus $E \rightarrowtail h_{l_m}$ for $m < i$ holds. Using (i) we get that for each argument $a \in A$ with $a \rightarrowtail r_{l_i}$, it holds that $E \rightarrowtail a$. Hence $r_{l_i} \in E$. Now in particular $r_{l_k} \in E$ and by (ii) we have that $E \rightarrowtail z$. As $z$ is the only argument attacking $t$ we also have that $t \in E$.

To show the if-part, let us assume that $t$ is contained in the grounded extensions $E$ of $F_{T,z}$. Then by construction $E \rightarrowtail z$ and thus there exists an integer $k$, such that $\mathcal{F}_F^k(\emptyset) \rightarrowtail z$ and for each $m < k : \mathcal{F}_F^m(\emptyset) \not\rightarrowtail z$. We claim that for $1 \leq m \leq k$ and $x \in X$ it holds that if $\mathcal{F}_F^m(\emptyset) \rightarrowtail x$ then $x$ is in the minimal model of $T$. The proof is by induction on $m$. As induction base consider $\mathcal{F}_F(\emptyset)$. By construction $\mathcal{F}_F(\emptyset)$ is the set of arguments that correspond to rules in $T$ having empty body. The arguments attacked by $\mathcal{F}_F(\emptyset)$ are the head atoms of these rules, which are clearly in the minimal model. For the induction step assume that $\mathcal{F}_F^{m-1}(\emptyset)$ only attacks arguments corresponding to atoms in the minimal model. As $\mathcal{F}_F^{m-1}(\emptyset) \not\rightarrowtail z$ we have that $t \notin \mathcal{F}_F^{m-1}(\emptyset)$. Let $x \in X$ be an argument such that $\mathcal{F}_F^m(\emptyset) \rightarrowtail x$, but $\mathcal{F}_F^{m-1}(\emptyset) \not\rightarrowtail z$. Then there exists an $r_i \in T$ such that $h_i = x$ and $r_i \in \mathcal{F}_F^m(\emptyset)$. By construction of $F_{T,z}$ we have that the argument $r_i$ is defended by $\mathcal{F}_F^{m-1}(\emptyset)$ iff each atom in the body of $r_i$ is attacked by $\mathcal{F}_F^{m-1}(\emptyset)$. Hence, by assumption each atom in the body of $r_i$ is contained in the minimal model of $T$. But then the head $h_i$ of $r_i$ is in the minimal model of $T$. Hence, as $\mathcal{F}_F^k(\emptyset) \rightarrowtail z$, we get that $z$ is in the minimal model of $T$. $\qquad\square$

**Proposition 2.** $\mathsf{Ver}_{stg}$ *is coNP-hard.*

*Proof.* We prove the assertion by reducing the (NP-hard) problem 3-SAT to the complementary problem of $\mathsf{Ver}_{stg}$. We assume that a 3-CNF formula is given as a set $C$ of clauses, where each clause is a set over atoms and negated atoms (denoted by $\bar{x}$). For such a CNF $\varphi$ over variables $X$, define the AF $F_\varphi = (A, R)$ with

$$
\begin{aligned}
A &= X \cup \bar{X} \cup C \cup \{s, t, b\} \\
R &= \{(x, \bar{x}), (\bar{x}, x) \mid x \in X\} \cup \{(l, c) \mid l \in c, c \in C\} \cup \\
&\quad \{(c, t) \mid c \in C\} \cup \{(s, y), (y, s) \mid y \in A \setminus \{s, b\}\} \cup \{(t, b), (b, b)\}
\end{aligned}
$$

where $\bar{X} = \{\bar{x} \mid x \in X\}$ and $s, t, b$ are fresh arguments. See Figure 3 for an illustrating example. We show that $\varphi$ is satisfiable iff $\{s\}$ is not a stage extension of $F_\varphi$. First let us assume $\varphi$ is satisfiable and let $T$ be any satisfying assignment of $\varphi$. Then the set $E = \{t\} \cup \{x \mid x \in X, T(x) = true\} \cup \{\bar{x} \mid x \in X, T(x) = false\}$ is a stable extension of $F_\varphi$, i.e. $E_R^+ = A$, and since $\{s\}_R^+ = A \setminus \{b\}$, $\{s\}$ is not a stage extension of $F_\varphi$. Now let us assume that $\{s\}$ is a stage extension. By the same argumentation as above, i.e. using $\{s\}_R^+ \subset A$, we get that $F_\varphi$ has no stable extension. But as we have seen before each satisfying assignment of $\varphi$ corresponds to a stable extension of $F_\varphi$. Thus we can conclude that $\varphi$ is unsatisfiable. $\qquad\square$

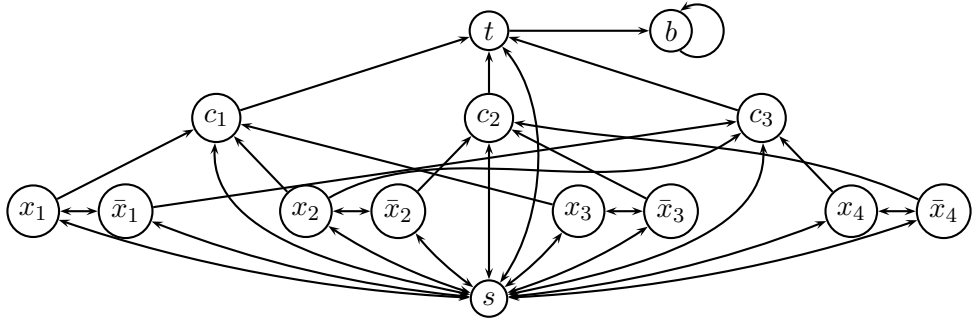

Figure 3: AF $F_{\{c_1,c_2,c_3\}}$ with $c_1 = \{x_1, x_2, x_3\}$, $c_2 = \{\bar{x}_2, \bar{x}_3, \bar{x}_4\}$, $c_3 = \{\bar{x}_1, x_2, x_4\}$.

Together with results from the literature (Coste-Marquis, Devred, & Marquis, 2005; Dimopoulos & Torres, 1996; Dung, 1995; Dunne & Bench-Capon, 2002; Dunne & Caminada, 2008; Dvořák & Woltran, 2010), we obtain the complexity-landscape of abstract argumentation as given in Table 1.

## 3. Properties for Translations

In what follows, we understand as a translation $Tr$ a function which maps AFs to AFs. In particular, we seek translations, such that for given semantics $\sigma, \sigma'$, the extensions $\sigma(F)$ are in a certain relation to extensions $\sigma'(F)$ for each AF $F$. To start with, we introduce a few additional properties which seem desirable for such translations. To this end, we define, for AFs $F = (A, R)$, $F' = (A', R')$, the union of AFs as $F \cup F' = (A \cup A', R \cup R')$, and inclusion as $F \subseteq F'$ iff jointly $A \subseteq A'$ and $R \subseteq R'$.

| $\sigma$ | $\mathsf{Cred}_\sigma$ | $\mathsf{Skept}_\sigma$ | $\mathsf{Ver}_\sigma$ | $\mathsf{Exists}_\sigma$ | $\mathsf{Exists}_\sigma^{\neg\emptyset}$ |
|---|---|---|---|---|---|
| $grd$ | P-c | P-c | P-c | trivial | in L |
| $stb$ | NP-c | coNP-c | in L | NP-c | NP-c |
| $adm$ | NP-c | trivial | in L | trivial | NP-c |
| $com$ | NP-c | P-c | in L | trivial | NP-c |
| $prf$ | NP-c | $\Pi_2^P$-c | coNP-c | trivial | NP-c |
| $sem$ | $\Sigma_2^P$-c | $\Pi_2^P$-c | coNP-c | trivial | NP-c |
| $stg$ | $\Sigma_2^P$-c | $\Pi_2^P$-c | coNP-c | trivial | in L |

Table 1: Complexity of abstract argumentation ($\mathcal{C}$-c denotes completeness for class $\mathcal{C}$).

**Definition 4.** *A translation Tr is called*

- efficient *if for every AF F, the AF Tr(F) can be computed using logarithmic space wrt. to* $|F|$;

- covering *if for every AF F, $F \subseteq Tr(F)$;*

- embedding *if for every AF F, $A_F \subseteq A_{Tr(F)}$ and $R_F = R_{Tr(F)} \cap (A_F \times A_F)$;*

- monotone *if for any AFs F,F', $F \subseteq F'$ implies $Tr(F) \subseteq Tr(F')$;*

- modular *if for any AFs F,F', $Tr(F) \cup Tr(F') = Tr(F \cup F')$.*

A translation should not reduce the expressiveness of a semantic using some expensive computation. Thus the computational cost of a translation should be less than the computational cost of any semantic under our focus, i.e. less than P. Thus using the class of logarithmic space computable functions is appropriate for our purposes. In addition, one could seek translations which are minimal wrt. certain parameters (for instance, number of additional arguments and attacks). However, we decided not to design our translations towards such aims, since this would partly hide the main intuitions underlying the translations.

While the property of efficiency is clearly motivated, let us spend a few words on the other properties. Covering holding ensures that the translation does not hide some original arguments or conflicts. Being embedding, in addition, ensures that no additional attacks between the original arguments are pretended. While efficiency is motivated by expressiveness and the possibility to reuse reasoning algorithms, the properties of covering and embedding can be motivated by the meta-argumentation scenario. Translations which are covering or embedding preserve the arguments and conflicts we (meta)-argue about, an assumption one usually has in mind in the context of meta-argumentation. To put it in other words, having an embedding translation, the original framework and the meta-level part are clearly separated in the translated framework.

Monotonicity and modularity are crucial when extending the source AF after translation. Let us first consider monotonicity. In multi-agent scenarios it may be impossible for one agent to withdraw already interchanged arguments and attacks, as the other agents may not agree to forget arguments and conflicts they already know about; hence, re-translating the augmented source AF should respect the already existing translation. Now let us consider modularity and adding only a few arguments/attacks to a huge AF. When updating the translation it suffices to only consider the new arguments/attacks, instead of the whole source AF, which indeed can be of computational value. In the field of meta-argumentation, modular translations are in particular interesting as they are compatible with merging AFs. Thus one can interchange merge- and translation-operations, i.e. it does not make a difference if one first merges two AFs and then translates the union or first translates both AFs and then merges the translations. Moreover, as it can be easily checked each modular transformation is also monotone.

Next, we give two properties which refer to semantics. We note that our concept of faithfulness follows the definition used by Janhunen (1999); while exactness is in the spirit of bijective faithfulness wrt. equivalence as used by Liberatore (2007).

**Definition 5.** *For semantics $\sigma, \sigma'$ we call a translation $Tr$*

- exact *for $\sigma \Rightarrow \sigma'$ if for every AF $F$, $\sigma(F) = \sigma'(Tr(F))$;*

- faithful *for $\sigma \Rightarrow \sigma'$ if for every AF $F$, $\sigma(F) = \{E \cap A_F \mid E \in \sigma'(Tr(F))\}$ and $|\sigma(F)| = |\sigma'(Tr(F))|$.*

However, due to the very nature of the different semantics we want to consider, we need some less restricted notions. For instance, if we consider a translation from stable to some other semantics, we have to face the fact that some AFs do not possess a stable extension, while other semantics always yield at least one extension. The following definition takes care of this issue.

**Definition 6.** *For semantics $\sigma, \sigma'$, we call a translation $Tr$*

- weakly exact *for $\sigma \Rightarrow \sigma'$ if there exists a collection $\mathcal{S}$ of sets of arguments, such that for any AF $F$, $\sigma(F) = \sigma'(Tr(F)) \setminus \mathcal{S}$;*

- weakly *faithful for $\sigma \Rightarrow \sigma'$ if there exists a collection $\mathcal{S}$ of sets of arguments, such that for any AF $F$, $\sigma(F) = \{E \cap A_F \mid E \in \sigma'(Tr(F)) \setminus \mathcal{S}\}$ and $|\sigma(F)| = |\sigma'(Tr(F)) \setminus \mathcal{S}|$.*

We sometimes refer to the elements from $\mathcal{S}$ as remainder sets. Note that $\mathcal{S}$ depends only on the translation, but not on the input AF. Thus, by definition, each $S \in \mathcal{S}$ only contains arguments which never occur in AFs subject to translation. In other words, we reserve certain arguments for introduction in weak translations.

Finally, we mention that the properties from Definition 4 as well as being exact, weakly exact and faithful are transitive, i.e. for two transformations satisfying one of these properties, also the concatenation satisfies the respective property. However, transitivity is not guaranteed for being weakly faithful.

## 4. Translations

In this section, we provide numerous faithful translations between the semantics introduced in Definition 3. As minimal desiderata, we want the translations to be efficient, monotone, and covering (see Definition 4). Thus, in this section when speaking about translations we tacitly assume that they satisfy at least these three properties.

### 4.1 Exact Translations

We start with a rather simple such translation, which we will show to be exact for $prf \Rightarrow sem$ and $adm \Rightarrow com$.

**Translation 1.** *The translation $Tr_1$ is defined as $Tr_1(F) = (A^*, R^*)$, where*

$$
\begin{aligned}
A^* &= A_F \cup A'_F \\
R^* &= R_F \cup \{(a, a'), (a', a), (a', a') \mid a \in A_F\},
\end{aligned}
$$

*with $A'_F = \{a' \mid a \in A_F\}$.*

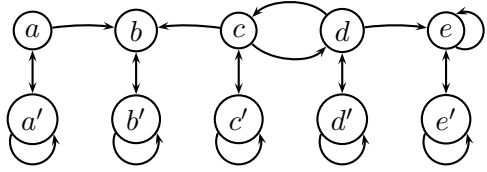

Figure 4: $Tr_1(F)$ for the AF $F$ from Example 1.

A few words about the intuition behind the above translation (for illustration see Figure 4 which depicts the translation of our example AF from Example 1): the new arguments $a' \in A'_F$ are all self-attacking and thus never appear in any extension of the resulting framework. However, each $a'$ attacks the original argument $a$ (and $a$ attacks $a'$), thus an argument $a$ is only defended by a set $E$ in $Tr_1(F)$ if $a \in E$. Consequently, we have that in $Tr_1(F)$ each admissible set is also a complete one.

**Lemma 1.** *For an AF $F$ and a set $E$ of arguments, the following propositions are equivalent:*

1. $E \in adm(F)$

2. $E \in adm(Tr_1(F))$

3. $E \in com(Tr_1(F))$

*Proof.* As all arguments in $A'_F$ are self-conflicting, every conflict-free set $E$ of $Tr_1(F)$ satisfies $E \subseteq A_F$. Further, since $Tr_1$ is embedding, $E$ is conflict-free in $F$ iff $E$ is conflict-free in $Tr_1(F)$. Moreover, since $Tr_1$ only adds symmetric attacks against arguments $a \in A_F$, we have that $E$ defends its arguments in $F$ iff $E$ defends its arguments in $Tr_1(F)$. Thus, $adm(F) = adm(Tr_1(F))$ and (1)⇔(2) follows. For (2)⇒(3), let $a \in A$ be an arbitrary argument and $E \subseteq A$. In $Tr_1(F)$ the argument $a$ is attacked by $a'$ and $a$ is the only attacker (except $a'$ itself) of $a'$. Hence, for each $a \in A$, $E$ defends $a$ only if $a \in E$ and thus every admissible set of $Tr_1(F)$ is also a complete one. Finally, (2)⇐(3) holds since $com(F) \subseteq adm(F)$ is true for any AF $F$. $\square$

Concerning $Tr_1$ we observe another side effect. As already mentioned $a \in A$ is the only argument attacking $a'$. Thus different preferred extensions of $Tr_1(F)$ have incomparable range (recall Definition 2), and therefore each preferred extension of $Tr_1(F)$ is also a semi-stable extension of $Tr_1(F)$.

**Lemma 2.** *For an AF $F$ and a set $E$ of arguments, the following propositions are equivalent:*

1. $E \in prf(F)$

2. $E \in prf(Tr_1(F))$

3. $E \in sem(Tr_1(F))$

*Proof.* For (1)⇔(2), it is sufficient to show that $E \in adm(F)$ iff $E \in adm(Tr_1(F))$ holds for each $E$. This is captured by Lemma 1. For (2)⇒(3), let $D, E \in prf(Tr_1(F))$ and, towards a contradiction, assume that $D^+_{R*} \subset E^+_{R*}$, i.e. $D \notin sem(Tr_1(F))$. As both $D$ and $E$ are preferred extensions, we have $D \not\subseteq E$. Thus, there exists an argument $a \in D \setminus E$. By construction of $Tr_1(F)$, we get $a' \in D_{R^*}^+$ but $a' \notin E_{R^*}^+$, a contradiction to $D_{R^*}^+ \subset E_{R^*}^+$. (2)⇐(3) follows from the fact $sem(F) \subseteq prf(F)$ for any AF $F$. □

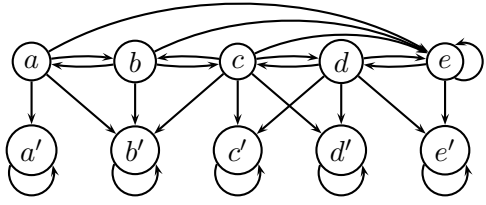

Figure 5: $Tr_2(F)$ for the AF $F$ from Example 1.

Obviously $Tr_1$ is an embedding translation and as the introduction of a new argument or attack in $Tr_1$ only depends on one original argument it is also modular. Together with the results from Lemma 1 and 2 we thus get our first main result.

**Theorem 1.** *$Tr_1$ is a modular, embedding, and exact translation for $prf \Rightarrow sem$ and $adm \Rightarrow com$.*

Our next translation, $Tr_2$, is concerned with stage and semi-stable semantics. In addition to $Tr_1$, we make all attacks from the original AF symmetric (thus $Tr_2$ will not be embedding) and add for each original attack $(a, b)$ also an attack $(a, b')$.

**Translation 2.** *The translation $Tr_2$ is defined as $Tr_2(F) = (A^*, R^*)$, where*

$$
\begin{aligned}
A^* &= A_F \cup A'_F \\
R^* &= R_F \cup \{(b, a), (a, b') \mid (a, b) \in R_F\} \\
&\quad \cup \{(a, b) \mid a \in A_F, (b, b) \in R_F\} \\
&\quad \cup \{(a, a'), (a', a') \mid a \in A_F\}
\end{aligned}
$$

The symmetric attacks in $Tr_2(F)$ mirror the fact that we do not mind the orientation of attacks when considering conflict-freeness. In other words, we exploit the well known property that for symmetric frameworks conflict-free and admissible sets coincide. However, making attacks symmetric destroys the original range of extensions. Thus we make use of arguments $a' \in A'_F$ in the sense that, for a given set $E$ of arguments, an argument $a'$ is contained in $E_{R^*}^+$ iff $a$ is contained in $E_R^+$. Likewise, we have to add attacks into self-defeating arguments. The technical reason for this is that we require that each original argument is attacked by a maximal conflict-free non-empty set in $Tr_2(F)$ (see also the proof of the forthcoming lemma). For illustration we refer to Figure 5.

**Lemma 3.** *For an AF $F$ and any set $E$ of arguments, the following propositions are equivalent:*

1. *$E \in stg(F)$*

2. *$E \in stg(Tr_2(F))$*

3. *$E \in sem(Tr_2(F))$*

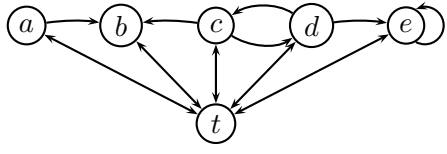

Figure 6: $Tr_3(F)$ for the AF $F$ from Example 1.

*Proof.* First, we mention that every stage extension of an AF $F$ is also maximal (wrt. $\subseteq$) conflict-free in $F$. Let us now consider the case where $\emptyset \in stg(F)$. We then have that $stg(F) = \{\emptyset\}$ which is equivalent to, for each $a \in A_F$ also $(a, a) \in R_F$. Then by construction of $Tr_2$ for each $a \in A^*$ also $(a, a) \in R^*$ and therefore $stg(Tr_2(F)) = sem(Tr_2(F)) = \{\emptyset\}$. Hence the lemma holds for such AFs, and for the remainder of the proof we can assume that $\emptyset \notin stg(F)$.

For (1)⇔(2), we again observe that a set $E$ is conflict-free in $F$ iff it is conflict-free in $Tr_2(F)$. In the following we use $(E_{R_F}^+)'$ as a short hand for $\{a' \in A' \mid a \in E_{R_F}^+\}$. Then we have that $(E_{R_F}^+)' \subseteq E_{R^*}^+$, since for each $(a, b) \in R_F$, we have $(a, b') \in R^*$. Furthermore, for each maximal conflict-free set $E$ in $F$ (and thus in $Tr_2(F)$), it holds that $A_F \subseteq E_{R^*}^+$. We show this by contradiction. To this end, let us assume that $A_F \not\subseteq E_{R^*}^+$, i.e. there exists $a \in A_F$ such that $a \notin E_{R^*}^+$. As $E \neq \emptyset$ we have that all self-attacking arguments are contained in $E_{R^*}^+$, thus $(a, a) \notin R^*$. As $a \notin E_{R^*}^+$ we have that $E \not\rightarrowtail^R a$ and $a \not\rightarrowtail^R E$, but then the set $E \cup \{a\}$ is conflict-free in $F$ and as $E$ is maximal $a \in E$; a contradiction. Hence, for each maximal conflict-free set $E \subseteq A_F$ in $F$, i.e. the candidates for stage extensions, it holds that $E_{R^*}^+ = A_F \cup (E_{R_F}^+)'$ and thus $E_{R_F}^+$ is maximal (wrt. subset inclusion) iff $E_{R^*}^+$ is maximal.

For (2)⇔(3), observe that each $a \in A_F$ with $(a, a) \notin R^*$ defends itself in $Tr_2(F)$ and all arguments $a' \in A_F'$ are self-conflicting. Thus, admissible and conflict-free sets coincide in $Tr_2(F)$. Consequently, the stage and semi-stable extensions of $Tr_2(F)$ coincide. □

By definition the translation $Tr_2$ is covering, but not embedding. Moreover, as each self-attacking argument is attacked by all of the other arguments $Tr_2$ is not modular. Together with the above lemma, we thus obtain the following result.

**Theorem 2.** *$Tr_2$ is an exact translation for stg ⇒ sem.*

The next translations consider the stable semantics as source formalism. Recall that not all AFs possess a stable extension, while this holds for all other semantics (also recall we excluded empty AFs for our considerations). Thus we have to use weak translations as introduced in Definition 6. Our first such translation is weakly exact and uses a single remainder set $\{t\}$ (recall the definition of remainder sets as given in Definition 6).

**Translation 3.** *The translation $Tr_3(F)$ is defined as $Tr_3(F) = (A^*, R^*)$ where*

$$A^* = A_F \cup \{t\}$$
$$R^* = R_F \cup \{(t, a), (a, t) \mid a \in A_F\}$$

Here the intuition is rather simple, see also Figure 6. In fact, the new argument $t$ in $Tr_3(F)$ encodes that there might not exist a stable extension for $F$. Thus none of the (other) arguments in $Tr_3(F)$ is accepted, whenever $t$ is accepted. Since the argument $t$ guards that there exists at least one stable extension of $Tr_3(F)$ (for any AF $F$), namely $\{t\}$, we can make use of the fact that stable, semi-stable and stage semantics thus coincide for $Tr_3(F)$.

**Lemma 4.** *Let $F = (A, R)$ be an AF and $E \subseteq A$. Then the following statements are equivalent:*

1. *$E \in stb(F)$*

2. *$E \in stb(Tr_3(F))$*

3. *$E \in sem(Tr_3(F))$*

4. *$E \in stg(Tr_3(F))$*

*Further for each $E \in \sigma(Tr_3(F))$ with $\sigma \in \{stb, sem, stg\}$ either $E = \{t\}$ or $t \notin E$ holds.*

*Proof.* As the translation does not modify the original AF $F$, i.e. $Tr_3$ is embedding, we have that for each $E \subseteq A_F$, $E$ is conflict-free in $F$ iff $E$ is conflict-free in $Tr_3(F)$.

(1)$\Rightarrow$(2): Each $E \in stb(F)$ by definition is non-empty, conflict-free and satisfies $E_{R_F}^+ = A_F$. By construction it also holds that $E \rightarrowtail^{R^*} t$ and thus $E_{R^*}^+ = A^*$, i.e. $E \in stb(Tr_3(F))$. For (1)$\Leftarrow$(2) consider $E \in stb(Tr_3(F)), E \subseteq A_F$. Then by definition we have that $E$ is conflict-free in $Tr_3(F)$) and thus in $F$; moreover, $E_{R^*}^+ = A^*$ and as $Tr_3$ is embedding also $E_R^+ = A_F$. Hence $E \in stb(F)$.

For (2)$\Leftrightarrow$(3)$\Leftrightarrow$(4), we mention that $\{t\}$ is a stable extension of $Tr_3(F)$ for any AF $F$. Furthermore, we know that if there exists a stable extension for an AF, then stable, semi-stable and stage extensions coincide.

Finally as the argument $t$ is in conflict with all of the other arguments the only extension $E$ with $t \in E$ is the set $\{t\}$. □

Adding argument $t$ and the corresponding attacks to the source AF is a modular operation and as no further attacks are added $Tr_3$ is also embedding.

**Theorem 3.** *$Tr_3$ is modular, embedding and weakly exact for $stb \Rightarrow \sigma$, $\sigma \in \{sem, stg\}$.*

*Proof.* The result follows from Lemma 4, which states that $sem(Tr_3(F)) = stg(Tr_3(F)) = stb(F) \cup \{\{t\}\}$. Thus by taking as remainder set $\mathcal{S} = \{\{t\}\}$, $Tr_3$ is weakly exact. □

We continue with a different translation from stable to other semantics.

**Translation 4.** *$Tr_4$ is defined as $Tr_4(F) = (A^*, R^*)$ where*

$$
\begin{aligned}
A^* &= A_F \cup A'_F \\
R^* &= R_F \cup \{(b', a) \mid a, b \in A_F\} \\
&\quad \cup \{(a', a'), (a, a') \mid a \in A_F\} \\
&\quad \cup \{(a, b') \mid (a, b) \in R_F\}.
\end{aligned}
$$

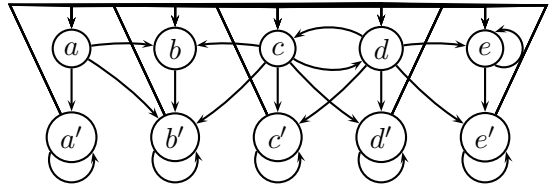

Figure 7: $Tr_4(F)$ for the AF $F$ from Example 1.

As before in translation $Tr_2$, new arguments $a' \in A'_F$ are used to encode the range of an extension in the sense that $a'$ is attacked by a set $E$ in $Tr_4(F)$ only if $a$ is in the range of $E$ in $F$. However, given the fact that each $a' \in A'_F$ attacks back all original arguments $a \in A$, we can now accept an argument in a set $E$ only if all arguments are in the range of $E$. For illustration on our running example, see Figure 7. Observe that in our example each of the arguments $a', b', c', d', e'$ attacks each of the arguments $a, b, c, d, e$.

**Lemma 5.** *Let $F = (A, R)$ be an AF and $E \subseteq A$ with $E \neq \emptyset$. Then, the following statements are equivalent:*

1. *$E \in stb(F)$*

2. *$E \in stb(Tr_4(F))$*

3. *$E \in adm(Tr_4(F))$*

4. *$E \in prf(Tr_4(F))$*

5. *$E \in com(Tr_4(F))$*

6. *$E \in sem(Tr_4(F))$*

*Further for each conflict-free set $E$ of $Tr_4(F)$ it holds that $E \subseteq A$.*

*Proof.* First, as all arguments $a' \in A'$ are self-attacking, for each conflict-free set $E$ in $Tr_4(F)$ it holds that $E \subseteq A$. Since the translation is embedding, any set $E$ is conflict-free in $F$ iff it is conflict-free in $Tr_4(F)$. To show (1)$\Rightarrow$(2), let $E \in stb(F)$. Hence, for all $a \in A \setminus E$, $E \rightarrowtail^R a$. We now claim that each argument in $A^* \setminus E$ is attacked by $E$ in $Tr_4(F)$. We distinguish between two cases for the different arguments in $A^* \setminus E$:

(i) $a \in A \setminus E$: The construction of $Tr_4(F)$ preserves all attacks in $R$. Thus as each $a \in A \setminus E$ satisfies $E \rightarrowtail^R a$, we obtain that $E \rightarrowtail^{R^*} a$

(ii) $a' \in A'$: In case $a \in E$ we have $E \rightarrowtail^{R^*} a'$, since $(a, a') \in R^*$ In case $a \in A \setminus E$, by the assumption $E \in stb(F)$, there exists an argument $b \in E$ such that $(b, a) \in R$. But then by construction $(b, a') \in R^*$ and thus $E \rightarrowtail^{R^*} a'$.

Together with our observations about conflict-free sets, we get $E \in stb(Tr_4(F))$.

Vice versa, to show (1)$\Leftarrow$(2) we get, for $E \in stb(Tr_4(F))$, $E \rightarrowtail^{R^*} a$, for each $a \in A^* \setminus E$, and thus, in particular, for each $a \in A \setminus E$. By definition of $Tr_4$, we also have $E \rightarrowtail^R a$ for each $a \in A \setminus E$. Thus $E \in stb(F)$ follows.

To show (2)⇐(3), let $E$ be a nonempty admissible extension of $Tr_4(F)$ and $a \in E$. By construction, we have that $a^- := \{b \in A^* \mid (b, a) \in R^*\} \supseteq A'$. As $E \in adm(Tr_4(F))$, $E \rightarrowtail^{R^*} a'$ for each $a' \in A'$. But $E \rightarrowtail a'$ only if either $a \in E$ or $E \rightarrowtail^{R^*} a$. Thus for every $a \in A^*$ it holds that either $a \in E$ or $E \rightarrowtail^{R^*} a$; hence, $E \in stb(Tr_4(F))$.

The remaining implications follow by well-known relations between the semantics, i.e. $stb(G) \subseteq sem(G) \subseteq prf(G) \subseteq com(G) \subseteq adm(G)$, for each AF $G$. Hence, in particular, since for $Tr_4(F)$, stable extensions and non-empty admissible sets coincide, the claim follows. □

Clearly $Tr_4$ is an embedding translation, but as for each new argument we add attacks to all original arguments, $Tr_4$ is not modular.

**Theorem 4.** *$Tr_4$ is an embedding and weakly exact translation for $stb \Rightarrow \sigma$ with $\sigma \in \{adm, com, prf, sem\}$.*

*Proof.* By Lemma 5, we in particular have that $stb(F) = \sigma(Tr_4(F)) \setminus \{\emptyset\}$, for any AF $F$. Thus taking $\emptyset$ as a remainder set, we obtain that $Tr_4$ is weakly exact for the involved semantics. □

Thus we have that both $Tr_3$ and $Tr_4$ are weakly exact translations for $stb \Rightarrow sem$, of course with different remainder sets. Due the to different properties of two translations it depends on the concrete application which of them would be the better choice.

## 4.2 Faithful Translations

So far, we have only introduced exact and weakly exact translations. We now present translations which relax this semantical property, i.e. we switch to faithful translations. As a first example, we consider a translation for $stg \Rightarrow sem$ which is faithful and embedding, but not exact. This is in contrast to translation $Tr_2$ which is exact for $stg \Rightarrow sem$ but not embedding. As we will see in Section 5 it is impossible to give a translation that is both embedding and exact for $stg \Rightarrow sem$, thus one has to decide which property is more important for a concrete application scenario.

**Translation 5.** *The translation $Tr_5(F)$ is defined as $Tr_5(F) = (A^*, R^*)$ where*

$$
\begin{aligned}
A^* &= A_F \cup \bar{A}_F \cup A'_F \\
R^* &= R_F \cup \{(a, \bar{a}), (\bar{a}, a) \mid a \in A_F\} \\
&\quad \cup \{(a, a'), (a', a') \mid a \in A_F\} \\
&\quad \cup \{(a, b') \mid (a, b) \in R_F\}
\end{aligned}
$$

As in $Tr_2(F)$ the arguments $a' \in A'_F$ handle the range of the original extensions. But instead of making original attacks symmetric (as in $Tr_2$) we add the arguments $\bar{a} \in \bar{A}_F$ to encode that an argument is not in the extension (also compare Figures 5 and 8). In fact, such meta-arguments indicating that some $a$ is out of an extension will be used in all faithful translations presented in this subsection.

**Lemma 6.** *Let $F = (A, R)$ be an AF, $E \subseteq A$ and $E^* = E \cup (\overline{A \setminus E})$. The following statements are equivalent:*

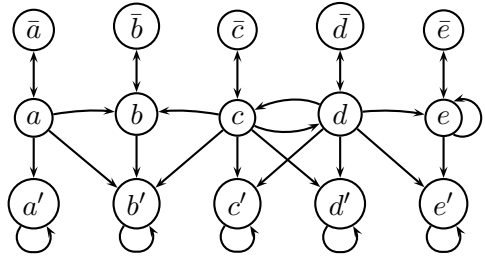

Figure 8: $Tr_5(F)$ for the AF $F$ from Example 1.

1. $E \in stg(F)$

2. $E^* \in stg(Tr_5(F))$

3. $E^* \in sem(Tr_5(F))$

*Moreover for each $S \in sem(Tr_5(F))$ there exists a set $E \subseteq A$ such that $S = E \cup \overline{(A \setminus E)}$.*

*Proof.* First we prove that each $S \in stg(Tr_5(F))$ is of the form $S = E \cup \overline{(A \setminus E)}$. As $S$ is conflict-free we have that $A'_F \cap S = \emptyset$ (each $a' \in A'$ is self-attacking) and for each $a \in A$ that $\{a, \bar{a}\} \not\subseteq E^*$ (as $a$ attacks $\bar{a}$ and vice versa). Further as each stage extension is also a $\subseteq$-maximal conflict-free set we have that for each $a \in A$ either $a \in S$ or $\bar{a} \in S$. Hence there exists an $E \subseteq A$ such that $S = E \cup \overline{(A \setminus E)}$.

(1)$\Rightarrow$(2): Let $E \in stg(F)$. It is easy to see that $E^*$ is conflict-free in $Tr_5(F)$. By construction for each argument $a \in A$ either $a \in E^*$ or $\bar{a} \in E^*$ holds and there are mutual attacks between $a$ and $\bar{a}$, hence we have that $A \cup \bar{A} \subseteq (E^*)^+_{R^*}$. Next we observe that each $a' \in A'$ is self-attacking and thus $a' \in (E^*)^+_{R^*}$ iff $E^* \rightarrowtail a'$. Further by the definition of $Tr_5(F)$ each argument $a'$ is attacked by $a$ and all arguments $b$ such that $(b, a) \in R$. That is $a' \in (E^*)^+_{R^*}$ iff either $a \in E$ or there exists a $b \in A$ such that $(b, a) \in R$ iff $a \in (E)^+_R$. By assumption $E$ is a stage extension of $F$ and thus we have that $(E)^+_R$ is $\subseteq$-maximal. Using the above observation we have that also $(E^*)^+_{R^*}$ is $\subseteq$-maximal in $Tr_5(F)$ and therefore $E^* \in stg(Tr_5(F))$.

(1)$\Leftarrow$(2): Let $E^* \in stg(Tr_5(F))$. We recall that $E^*$ is of the form $S = E \cup \overline{(A \setminus E)}$, for some $E \subseteq A$. It can be easily checked that $E$ is conflict-free in $F$. By the above observation that $a' \in (E^*)^+_{R^*}$ iff $a \in (E)^+_R$ and the fact that $(E^*)^+_{R^*}$ is $\subseteq$-maximal in $Tr_5(F)$ we get that also $E^+_R$ is $\subseteq$-maximal in $F$. Hence, $E \in stg(F)$.

(2)$\Leftrightarrow$(3): Let us consider $E^* \in stg(Tr_5(F))$. As we have already observed, $E^*$ is of the desired form and for each $a \in A_F \cup \bar{A}_F$ either $a \in E^*$ or $E^* \rightarrowtail a$. Further by construction an argument $b \in A'_F$ does not attack $E^*$. We can conclude that each stage extension defends itself against all attackers, i.e. is an admissible set. Hence, stage and semi-stable extensions of $Tr_5(F)$ coincide. $\square$

By above lemma and construction of $Tr_5$, the following result is immediate.

**Theorem 5.** *$Tr_5$ is a modular, embedding and faithful translation for $stg \Rightarrow sem$.*

Next we give a faithful translation from admissible semantics to stable, semi-stable and stage semantics.

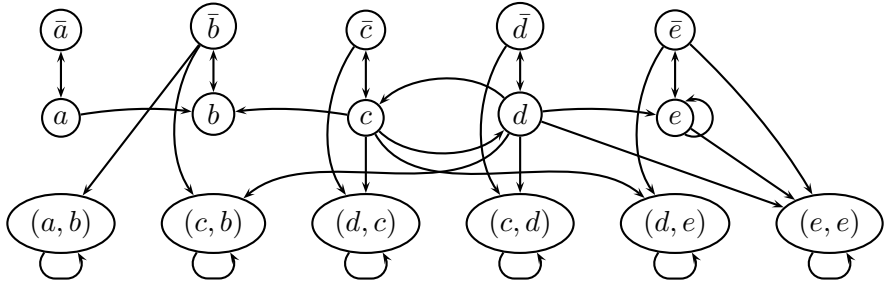

Figure 9: $Tr_6(F)$ for the AF $F$ from Example 1.

**Translation 6.** *The translation $Tr_6(F)$ is defined as $Tr_6(F) = (A^*, R^*)$ where*

$$
\begin{aligned}
A^* &= A_F \cup \bar{A}_F \cup R_F \\
R^* &= R_F \cup \{(a, \bar{a}), (\bar{a}, a) \mid a \in A_F\} \\
&\quad \cup \{(r, r) \mid r \in R_F\} \\
&\quad \cup \{(\bar{a}, r) \mid r = (y, a) \in R_F\} \\
&\quad \cup \{(a, r) \mid r = (z, y) \in R_F, (a, z) \in R_F\}
\end{aligned}
$$

The main idea is to use additional arguments $(a, b) \in A^*$ which represent the attack relations from the source framework in order to capture admissibility as follows: $(a, b)$ is attacked by an extension $E^*$ in $Tr_6(F)$ if $(a, b)$ is not critical wrt. the corresponding extension $E$ in $F$, meaning that either $b \notin E$ or there exists a $c \in E$ such $(c, b) \in R_F$, i.e. $a$ is defended by $E$. For instance, consider the argument $(c, b)$ in the translation of our example framework as depicted in Figure 9. Then, we have that (1) $\bar{b}$ attacks $(c, b)$ since if $b$ is chosen to be out (i.e. $\bar{b}$ is chosen in), there is no need to defend $b$; (2) $d$ attacks $(c, b)$ since if $d$ is chosen in, $d$ defends $b$ against attacker $c$ (recall that $(d, c)$ is present in the source AF). Thus, as long as $(c, b)$ is attacked by some argument, $b$ is treated corrected in terms of admissibility (wrt. attacker $c$). Note that in our example $b$ cannot be defended against $a$, thus the only way to get $(a, b)$ into the range is to select $b$ to be out.

**Lemma 7.** *Let $F = (A, R)$ be an AF, $E \subseteq A$ and $E^* = E \cup (\overline{A \setminus E})$. The following statements are equivalent:*

1. $E \in adm(F)$

2. $E^* \in stb(Tr_6(F))$

3. $E^* \in sem(Tr_6(F))$

4. $E^* \in stg(Tr_6(F))$

*Moreover for each $E^* \in \sigma(Tr_6(F))$ $(\sigma \in \{stb, sem, stg\})$ there exists a set $E \subseteq A$ such that $E^* = E \cup (\overline{A \setminus E})$.*

*Proof.* $(1) \Rightarrow (2)$: Let $E \in adm(F)$. It is easy to see that $E^*$ is conflict-free in $Tr_6(F)$ and further that $A \cup \bar{A} \subseteq (E^*)^+_{R^*}$. It remains to show that each argument $r \in A^*$ for $r \in R$ is attacked by $E^*$. Let $(a, b)$ be such an argument $r$. If $b \notin E$ then $\bar{b} \in E^*$ and thus $E^* \rightarrowtail^{R^*} r$. Otherwise, $b \in E$ (thus $b \in E^*$) and, by assumption, $E$ defends $b$ in $F$, i.e. $(c, a) \in R$ for some $c \in E$ (thus $c \in E^*$). By construction, $(c, r) \in R^*$ and $E^* \rightarrowtail^{R^*} r$.

(1)$\Leftarrow$(2): Let $E^* \in stb(Tr_6(F))$. $E^*$ is conflict-free, thus $R \cap E^* = \emptyset$ and $\{a, \bar{a}\} \nsubseteq E^*$ for all $a \in A$. By construction, $E$ is conflict-free in $F$. It remains to show that $E$ defends all its arguments in $F$. Let $b \in A \setminus E$ such that $b \rightarrowtail^R a$ for some $a \in E$. Then there exists an argument $(b, a)$ in $Tr_6(F)$ attacked by $E$. As $a \in E$ we have that $\bar{a} \notin E^*$ and thus there exists an argument $c \in E$ such that $(c, b) \in R$.

(2)$\Leftrightarrow$(3)$\Leftrightarrow$(4): As the empty set is always admissible we have that $\bar{A}$ is always a stable extension of $Tr_6(F)$. Hence, stable, semi-stable and stage extensions coincide in $Tr_6(F)$, for any AF $F$. □

Observe that in the construction of $Tr_6$ drawing attacks $\{(a, r) \mid r = (z, y) \in R_F, (a, z) \in R_F\}$ depends on two attacks and three arguments from the original framework. Hence $Tr_6$ is not modular. By Lemma 7 the next result follows quite easily.

**Theorem 6.** *Translation $Tr_6$ is embedding and faithful for $adm \Rightarrow \sigma$ ($\sigma \in \{stb, sem, stg\}$).*

In our faithful translation from complete to stable semantics which we present next, we extend the given AF by arguments that represent whether an argument is attacked in the corresponding extension or not. Further we add arguments that ensure admissibility and completeness. The entire translation is thus slightly more complicated; see also Figure 10 which depicts the translated framework for our running example.

**Translation 7.** *The translation $Tr_7(F)$ is defined as $Tr_7(F) = (A^*, R^*)$ where*

$$A^* = A_F \cup \bar{A}_F \cup A_F^\circ \cup \bar{A}_F^\circ \cup A_F' \cup R_F$$
$$R^* = R_F \cup \{(x, x) \mid x \in A_F' \cup R_F\}$$
$$\cup \{(a, \bar{a}), (\bar{a}, a), (\bar{a}^\circ, a^\circ), (a, a') \mid a \in A_F\}$$
$$\cup \{(a, \bar{b}^\circ), (\bar{a}^\circ, b') \mid (a, b) \in R_F\}$$
$$\cup \{(\bar{a}, r'), (b^\circ, r') \mid r = (b, a) \in R_F\}$$

The intuition behind arguments $A_F'$, $\bar{A}_F$, and $R_F$ is similar as in previous translations. An argument $a^\circ \in A_F^\circ$ indicates that $a$ is attacked by an extension $E$ of $F$, while $\bar{a}^\circ \in \bar{A}_F^\circ$ says that $a$ is not attacked by $E$.

**Lemma 8.** *Let $F = (A, R)$ be an AF, $E \subseteq A$ and $E^* = E \cup \overline{(A \setminus E)} \cup \{a^\circ \mid E \rightarrowtail^R a\} \cup \{\bar{a}^\circ \mid E \not\rightarrowtail^R a\}$. Then the following statements are equivalent:*

1. $E \in com(F)$

2. $E^* \in stb(Tr_7(F))$

3. $E^* \in sem(Tr_7(F))$

4. $E^* \in stg(Tr_7(F))$

*Moreover for each $E^* \in \sigma(Tr_6(F))$ ($\sigma \in \{stb, sem, stg\}$) there exists a set $E \subseteq A$ such that $E^* = E \cup \overline{(A \setminus E)} \cup \{a^\circ \mid E \rightarrowtail^R a\} \cup \{\bar{a}^\circ \mid E \not\rightarrowtail^R a\}$.*

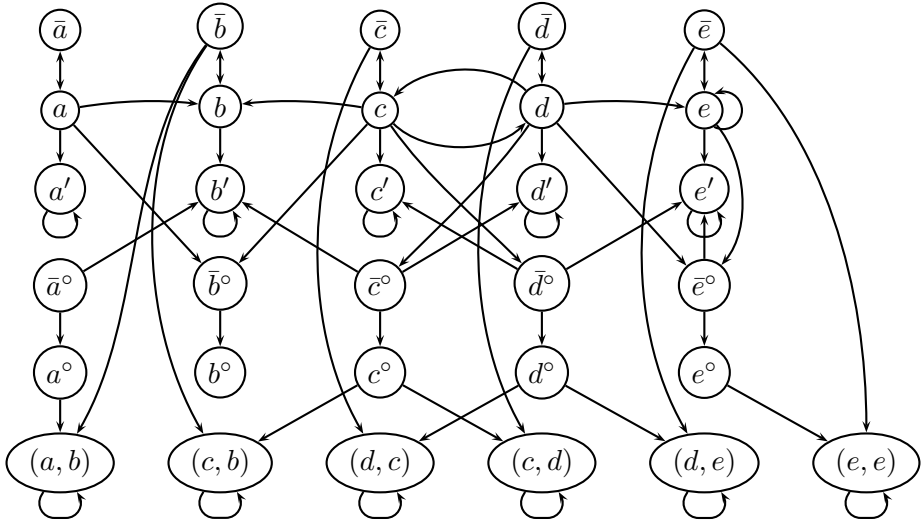

Figure 10: $Tr_7(F)$ for the AF $F$ from Example 1.

*Proof.* To show (1)⇒(2), let $E \in com(F)$. Then by construction $E^*$ is conflict-free in $Tr_7(F)$ (for $x, y \in E$ we have $x \rightarrowtail^R y \Leftrightarrow x \rightarrowtail^{R^*} y$). Moreover, by definition of $E^*$, it can be verified that $A \cup \bar{A} \cup A^\circ \cup \bar{A}^\circ \subseteq (E^*)_{R^*}^+$. Thus it remains to show that (i) $A' \subseteq (E^*)_{R^*}^+$ and (ii) $R \subseteq (E^*)_{R^*}^+$.

(i) Let $a \in A$ be an arbitrary argument of $F$. As $E$ is a complete extension we have that either $a \in E$, and thus $a \in E^*$, or there exists an attack $(b, a) \in R$ with $E \not\rightarrowtail^R b$, and thus $\bar{b}^\circ \in E^*$. As by construction $(\bar{b}^\circ, a') \in R^*$ we thus have that $E^* \rightarrowtail^{R^*} a'$.

(ii) Let $r = (b, a) \in R$ be an arbitrary attack of $F$. As $E$ is admissible it holds that either $a \notin E$, and thus $\bar{a} \in E^*$, or $E \rightarrowtail^R b$, and thus $b^\circ \in E^*$. In both cases $E^* \rightarrowtail^{R^*} r$.

Putting things together, we get that $A \cup \bar{A} \cup A^\circ \cup \bar{A}^\circ \cup A' \cup R = A^* \subseteq (E^*)_{R^*}^+$ which is equivalent to $E^*$ being a stable extension of $Tr_7(F)$.

To show (1)⇐(2), let $E^* \in stb(Tr_7(F))$. First we prove that $E^*$ is of the desired form. As $E^*$ is both conflict-free and $\subseteq$-maximal we clearly have that $E^* \cap (A \cup \bar{A}) = E \cup \overline{A \setminus E}$ for some $E \subseteq A$. Let now $a \in A$ be an arbitrary argument. We have that $a^\circ \in E^*$ iff $\bar{a}^\circ \notin E^*$. But as $E^*$ is stable $\bar{a}^\circ \notin E^*$ iff there exists an attack $(b, \bar{a}^\circ)$ such that $b \in E^*$. By construction of $Tr_7(F)$ this is equivalent to $b \in E$ and therefore $E \rightarrowtail^R a$. Thus $E^*$ is of the desired form, it remains to show that $E$ is complete. As mentioned before we have for $x, y \in E : x \rightarrowtail^R y \Leftrightarrow x \rightarrowtail^{R^*} y$ and thus $E$ is conflict-free in $F$. Thus it remains to show that (i) $E$ defends each of its arguments in $F$ and (ii) $E$ contains each argument defended by $E$ in $F$.

(i) Let us assume there exists an argument $a \in E$ not defended by $E$. Thus there exists $r = (b, a) \in R, E \not\rightarrowtail b$. By construction we also have that $\bar{a} \notin E^*$ (as $a \in E$) and $b^\circ \notin E^*$ (as $E \not\rightarrowtail b$). But in $Tr_7(F)$ the self-attacking argument $r$ is only attacked by the arguments $\bar{a}, b^\circ$ (and itself). Hence, this is in contradiction to $E^*$ being a stable extension.

(ii) Let $a \in A$ be an argument defended by $E$. Then for all arguments $b \rightarrowtail^R a$ we have that $E \rightarrowtail^R b$ and thus $b^\circ \in E^*$ and $\bar{b}^\circ \notin E^*$. Recall that in $Tr_7(F)$ the argument $a'$ is self-attacking and thus does not belong to $E^*$ and is only attacked by the arguments $\bar{b}^\circ$ and $a$. As $E^*$ is a stable extension and $a' \notin E^*$ we have that $a \in E^*$ and $a \in E$.

$(2) \Leftrightarrow (3) \Leftrightarrow (4)$: As there always exists a complete extension we know that any framework $Tr_7(F)$ has a stable extension. But then stable, stage and semi-stable extensions coincide. □

Translation $Tr_7$ introduces a huge number of new arguments, despite this the introduction of a concrete argument or attack only depends on a single argument or attack. Hence $Tr_7$ is modular. It is easily checked that $Tr_7$ is also embedding. Together with Lemma 8 we thus can state the following result for $Tr_7$.

**Theorem 7.** *$Tr_7$ is a modular, embedding and faithful translation for com $\Rightarrow \sigma$ ($\sigma \in \{stb, sem, stg\}$).*

Finally we present a translation from grounded semantics to most of the other semantics under our focus, i.e. to all semantics except admissible semantics. The main idea is to simulate the computation of the least fixed-point of the characteristic function $\mathcal{F}_F(S) = \{x \in A_F \mid x \text{ is defended by } S\}$ of an AF $F$ within the target AF.

**Translation 8.** *The translation $Tr_8(F)$ is defined as $Tr_8(F) = (A^*, R^*)$ where*

$$\begin{aligned} A^* &= A_{F,1} \cup \bar{A}^\circ_{F,1} \cup \cdots \cup A_{F,l} \cup \bar{A}^\circ_{F,l} \\ R^* &= R_F \cup \{(\bar{a}^\circ_i, b_i) \mid (a,b) \in R, i \in [l]\} \\ &\quad \cup \{(a_i, \bar{b}^\circ_{i+1}) \mid (a,b) \in R, i \in [l-1]\} \end{aligned}$$

*with $A_F = A_{F,l}$ and $l = \lceil \frac{|A_F|}{2} \rceil$.*

For illustration, we use here a slightly different example depicted in Figure 11(a). Observe that this AF has $\{a, c, d\}$ as its grounded extension. The translated framework is given in Figure 11(b).

The intuition behind arguments $a_i \in A_{F,i}$ is that $a \in \mathcal{F}^i_F(\emptyset)$, while the intuition of $\bar{a}^\circ_i \in \bar{A}^\circ_{F,i}$ is that $\mathcal{F}^{(i-1)}_F(\emptyset) \not\rightarrow a$. The integer $l$ is an upper bound for the number of iterations we need to reach the least fixed-point, i.e. the grounded extension.

**Lemma 9.** *Let $F = (A, R)$ be an AF and $E^*$ the grounded extension of $Tr_8(F)$. Then $E^* \cap A$ is the grounded extension of $F$. We further have that on $Tr_8(F)$ the grounded, stable, complete, preferred, semi-stable and stage extensions coincide.*

*Proof.* We recall the definition of the characteristic function $\mathcal{F}_F$ of an AF $F$, defined as $\mathcal{F}_F(S) = \{x \in A_F \mid x \text{ is defended by } S\}$, and that the grounded extension of $F$ is the least fix-point of $\mathcal{F}_F$. Further we use as a shorthand $F^* = Tr_8(F)$. One can show that for arbitrary $a \in A$ we have

(i) $a_i \in E^*$ iff $a \in \mathcal{F}^i_F(\emptyset)$;

(ii) $\bar{a}^\circ_i \in E^*$ iff $\mathcal{F}^{i-1}_F(\emptyset) \not\rightarrow^R a$; and

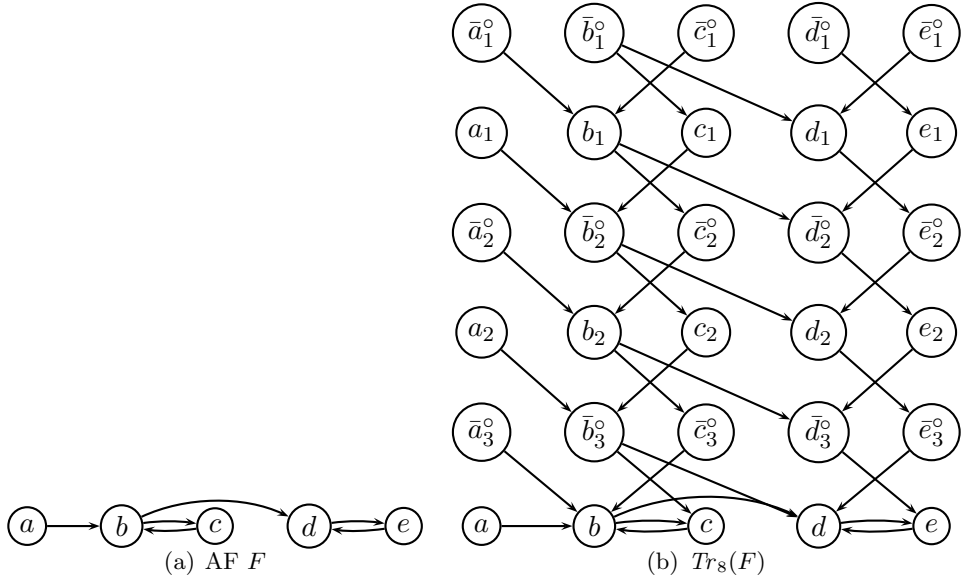

Figure 11: An example for $Tr_8$.

(iii) $A_{F,i} \subseteq (E^*)^+_{R^*}$.

(iv) $\bar{A}^\circ_{F,i} \subseteq (E^*)^+_{R^*}$.

We prove this by structural induction. As induction base we show (ii) and (iv) for the arguments $\bar{a}^\circ_1$. For $a \in A$ we have that $\bar{a}^\circ_1 \in E^*$ as they are not attacked by any argument. This coincides with the fact that $\mathcal{F}^0_F(\emptyset) = \emptyset$ doesn't attack any argument and thus (ii) and (iv) holds.

We have two induction steps: (1) Showing that (i) and (iii) hold for arbitrary $n$ iff (ii) and (iv) hold for $n$; and (2) showing that (ii) and (iv) hold for arbitrary $n$ iff (i) and (iii) hold for $n - 1$.

(1) We assume (ii) and (iv) hold for all $\bar{a}^\circ_n$. By the definition of $\mathcal{F}_F$ we have that $a \in \mathcal{F}^n_F(\emptyset)$ iff all $b \in a^- = \{b \in A \mid b \rightarrowtail a\}$ are attacked by $\mathcal{F}^{n-1}_F(\emptyset)$. Applying the induction hypothesis (ii) to $b \in a^-$ we obtain that $a \in \mathcal{F}^n_F(\emptyset)$ iff each $\bar{b}^\circ_i \in \{\bar{b}^\circ_i \mid (b,a) \in R\}$ is attacked by $E^*$. Further, as by the construction of $Tr_8(F)$ these are the only attackers of $a$, this is equivalent to argument $a_i$ being defended by $E^*$. Now recall that the each argument defended by the grounded extension is indeed contained in the grounded extension. Hence, $a \in \mathcal{F}^n_F(\emptyset)$ iff $a_i \in E^*$ and (i) holds.

To show (iii) we consider $a_i \in A_{F,i}$. If $a_i \in E^*$ then clearly $a_i \in (E^*)^+_{R^*}$. Thus let us consider $a_i \notin E^*$. Then, by the above observations, there exists a $\bar{b}^\circ_i$ such that $\bar{b}^\circ_i \rightarrowtail a_i$ and $E^* \not\rightarrowtail \bar{b}^\circ_i$. Using the latter and the induction hypothesis (iv) we obtain that $\bar{b}^\circ_i \in E^*$. Now we have that $E^* \rightarrowtail a_i$, hence $a_i \in (E^*)^+_{R^*}$ and we obtain (iii).

(2) Now let us assume that (i) and (iii) hold for all $a_{n-1}$. We have that $\mathcal{F}^{n-1}_F(\emptyset) \rightarrowtail a$ iff there exists $b \in \mathcal{F}^{n-1}_F(\emptyset) \cap \{b \mid (b,a) \in R^*\}$. By induction hypothesis this holds iff there exists a $b_{i-1} \in E^*$ such that $(b,a) \in R$. In other words there exists $b_{i-1} \in E^*$ such that $b_{i-1} \rightarrowtail^{R^*} \bar{a}_i^\circ$, which implies that $\bar{a}_i^\circ \notin E^*$. Moreover if there is no $b_{i-1} \in E^*$ such that $b_{i-1} \rightarrowtail^{R^*} \bar{a}_i^\circ$, by assumption (iii) we have that $E^*$ defends $\bar{a}_i^\circ$ and thus $\bar{a}_i^\circ \in E^*$. Hence (ii) and (iv) hold.

Furthermore when applying the $\mathcal{F}_F$ operator we either add a new argument to the set and attack an additional argument or we reach the fixed-point. So in each step we make a decision about at least two arguments and thus $\mathcal{F}_F^l(\emptyset) = grd(F)$. In combination with (i), we get that $a_l \in E^*$ iff $a \in grd(F)$. Moreover by (iii) and (iv) it holds that $E^*$ is also a stable extension and thus $grd(F^*) = stb(F^*) = com(F^*) = prf(F^*) = sem(F^*) = stg(F^*)$.  □

As in $Tr_8$ the integer value $l$ depends on the size $S$ of the source AF, $Tr_8$ is not modular. However, it can be verified that the computation of the translation only requires logarithmic space wrt. $S$ and that $Tr_8$ is embedding (the original AF is indeed contained in the resulting AF; see also the bottom layer in Figure 11(b)). Our final result concerning translations thus follows immediately from Lemma 9.

**Theorem 8.** *$Tr_8$ is an embedding and faithful translation for $grd \Rightarrow \sigma$ ($\sigma \in \{stb, com, prf, stg, sem\}$).*

## 5. Negative Results

In this section, we present results fortifying that for several semantics there does not exist any translation with the desired properties. The first result, which is rather straight forward, relies on the fact that the grounded semantics is a unique-status semantics.

**Proposition 3.** *There is no (weakly) faithful translation for $\sigma \Rightarrow grd$ with $\sigma \in \{sem, stg, prf, com, stb, adm\}$.*

*Proof.* For instance consider the AF $F = (\{a,b\}, \{(a,b),(b,a)\})$. We have that $\{\{a\}, \{b\}\} \subseteq \sigma(F)$ for $\sigma \in \{sem, stg, prf, com, stb, adm\}$ but the grounded semantics always proposes a unique extension.  □

We observe that in general it holds that if $\sigma$ is a multiple status semantics and $\sigma'$ is a unique status semantics then there is no (weakly) faithful translation for $\sigma \Rightarrow \sigma'$.

Further results are based on complexity gaps between different semantics (see Table 1) and the fact that certain translations preserve some decision problem. We start with cases where it is impossible to find efficient faithful translations; even if we allow for weakly faithful translations, cf. Definition 6. Afterwards, we give some negative results concerning (weakly) exact translations.

The following theorem concerns the intertranslatability of preferred, semi-stable and stage semantics, i.e. the semantics where skeptical acceptance is $\Pi_2^P$-complete. The underlying reason for the impossibility result is the complexity gap for the credulous acceptance problems.

**Theorem 9.** *There is no efficient (weakly) faithful translation for $sem \Rightarrow prf$ or $stg \Rightarrow prf$ unless $\Sigma_2^P = NP$.*

*Proof.* Let $Tr$ be an efficient (weakly) faithful translation from $\sigma \in \{sem, stg\}$ to $prf$. By definition this translation is L-computable and as we show next reduces $\mathsf{Cred}_\sigma$ to $\mathsf{Cred}_{prf}$: Let $F = (A, R)$ be an arbitrary AF, $x \in A$ an argument. First let us assume that $x$ is credulously accepted wrt. to $\sigma$. Hence, there exists an $E \in \sigma(F)$ with $x \in E$. As $Tr$ is a weakly faithful translation, there is an $E^* \in prf(Tr(F))$, such that $E^* \cap A = E$. Thus $x \in E^*$, i.e. $x$ is credulously accepted wrt. preferred semantics in $Tr(F)$.

So assume $x$ is credulously accepted in $Tr(F)$ wrt. to $prf$, i.e. $x \in E^*$ for some $E^* \in prf(Tr(F))$. By $x \in E^* \cap A$ we can conclude that $E^*$ is not a remainder set of $Tr$. As $Tr$ is a weakly faithful translation we have that $E = E^* \cap A$ is in $\sigma(F)$, and thus $x$ is credulously accepted in $F$ wrt. $\sigma$. Thus, $Tr$ is a L-reduction from the $\Sigma_2^P$-hard problem $\mathsf{Cred}_\sigma$ to the NP-easy problem $\mathsf{Cred}_{prf}$. □

The following theorem makes use of complexity gaps for the skeptical acceptance.

**Theorem 10.** *There is no efficient (weakly) faithful translation for $\sigma \Rightarrow \sigma'$, where $\sigma \in \{sem, stg, prf\}$ and $\sigma' \in \{com, stb, adm\}$, unless $\Sigma_2^P = \mathrm{NP}$.*

*Proof.* Given an efficient weakly faithful translation $Tr$ with remainder set $\mathcal{S}$ for $\sigma \Rightarrow \sigma'$ we have that $\mathsf{Skept}_\sigma$ is translated to the problem $\mathsf{Skept}_{\sigma'}^{\mathcal{S}}$, that is deciding whether an argument is in each $\sigma'$-extension which is not in the set $\mathcal{S}$. Next we show that the problem $\mathsf{Skept}_{\sigma'}^{\mathcal{S}}$ remains in coNP. One can disprove $\mathsf{Skept}_{\sigma'}^{\mathcal{S}}$, by guessing a set $E \subseteq A$, such that $a \notin E$ and verify that $E \in \sigma'(F)$ and $E \notin \mathcal{S}$. As $\mathsf{Ver}_{\sigma'} \in \mathrm{P}$ and the set $\mathcal{S}$ is fixed, i.e. $\mathcal{S}$ does not depend on the input, this is an NP-algorithm. Hence proving $\mathsf{Skept}_{\sigma'}^{\mathcal{S}}$ is in coNP. Thus $Tr$ would be an L-reduction from the $\Pi_2^P$-hard problem $\mathsf{Skept}_\sigma$ to the coNP-easy problem $\mathsf{Skept}_{\sigma'}^{\mathcal{S}}$, which implies $\Sigma_2^P = \mathrm{NP}$. □

One might prefer (weakly) exact over (weakly) faithful translations. As we have seen in Section 4, several of our translations are not exact but only faithful. In these cases we are interested in either finding an exact translation or an evidence that an exact translation is not possible. The following theorems approve that it was appropriate to have given only a (weakly) faithful translation in Section 4, as there cannot be any exact such translation.

**Theorem 11.** *There is no (weakly) exact translation for $\sigma \Rightarrow \sigma'$ where $\sigma \in \{adm, com\}$ and $\sigma' \in \{stb, prf, sem, stg\}$.*

*Proof.* This is basically by the fact that admissible resp. complete extensions may be in a $\subset$-relation; consider e.g. $F = (\{a, b\}, \{(a, b), (b, a)\})$ with $\sigma(F) = \{\{a\}, \{b\}, \emptyset\}$. Let us now assume there exists a (weakly) exact translation $Tr$ for $\sigma \Rightarrow \sigma'$. By definition, $\sigma(F) = \{\{a\}, \{b\}, \emptyset\} \subseteq \sigma'(Tr(F))$, but as $\emptyset \subset \{a\}$ this contradicts $\sigma' \in \{stb, prf, sem, stg\}$. □

**Theorem 12.** *There is no (weakly) exact translation for $com \Rightarrow adm$.*

*Proof.* We observe that for every AF $F$ it holds that $\emptyset \in adm(F)$, but there are AFs where $\emptyset \notin com(F)$. Thus for a weakly exact translation $Tr$, with the collection $\mathcal{S}$ of remainder sets, it holds that $\emptyset \in \mathcal{S}$. But then, given an AF $F$ with $\emptyset \in com(F)$, e.g. $F = (\{a, b\}, \{(a, b), (b, a)\})$, we can conclude that $\emptyset \in adm(Tr(F)) \setminus \mathcal{S}$, a contradiction. □

**Theorem 13.** *There is no efficient (weakly) exact translation for $grd \Rightarrow \sigma$ where $\sigma \in \{stb, adm, com\}$, unless $\mathrm{L} = \mathrm{P}$.*

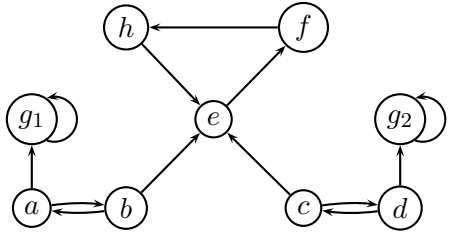

Figure 12: Counterexample for exact translations $\sigma \Rightarrow stg$ ($\sigma \in \{sem, prf\}$).

*Proof.* Let us, towards a contradiction, assume that there exists an efficient (weakly) exact translation $Tr$ for $grd \Rightarrow \sigma$. For a given AF $F = (A, R)$ with a set $E \subseteq A$ it holds that $E \in grd(F)$ iff $E \in \sigma(Tr(F))$. Thus $Tr$ would be an L-reduction from the P-hard problem $\mathsf{Ver}_{grd}$ (see Proposition 1) to $\mathsf{Ver}_{\sigma}$ ($\sigma \in \{stb, adm, com\}$) which is in L. $\qquad \square$

In Section 4 we presented two translations for $stg \Rightarrow sem$: $Tr_2$ which is an exact translation, but not embedding, and $Tr_5$ which is an embedding and faithful translation, but not exact. Let us also mention at this point that $Tr_2$ was the only translation presented in Section 4 that is not embedding. Hence a natural question that occurs is whether a translation that is embedding and exact for $stg \Rightarrow sem$ is possible. We give a negative answer to this question.

**Theorem 14.** *There is no embedding and (weakly) exact translation for $stg \Rightarrow sem$.*

*Proof.* Let us assume there exists an embedding and (weakly) exact translation $Tr$ for $stg \Rightarrow sem$. Consider the AF $F = (\{a, b\}, \{(a, a), (a, b)\})$ with $stg(F) = \{\{b\}\}$. As $Tr$ is a (weakly) exact translation we have that $\{b\} \in sem(Tr(F))$ and thus $\{b\} \in adm(Tr(F))$. Further we have that $(a, b) \in R_{Tr(F)}$ ($Tr(F)$ is embedding) and thus $\{b\}$ must attack $a$. But then we have $(b, a) \in R_{Tr(F)}$ which is contradiction to $Tr$ being an embedding translation. $\qquad \square$

Finally we present an impossibility result for $prf \Rightarrow stg$ and $sem \Rightarrow stg$.

**Theorem 15.** *There is no (weakly) exact translation for $\sigma \Rightarrow stg$ ($\sigma \in \{sem, prf\}$).*

*Proof.* Consider the AF $F = (\{a, b, c, d, e, f, g_1, g_2, h\}, \{(g_1, g_1), (g_2, g_2), (a, b), (b, a), (c, d), (d, c), (a, g_1), (b, e), (c, e), (d, g_2), (e, f), (f, h), (h, e)\})$ illustrated in Figure 12. We have that $sem(F) = \{\{b, d, f\}, \{a, c, f\}, \{a, d\}\}$ and $prf(F) = sem(F) \cup \{\{b, c, f\}\}$.

To prove that there is no weakly exact translation for $\sigma \Rightarrow stg$ ($\sigma \in \{sem, prf\}$), we will show that there exists no AF $F'$ with $sem(F) \subseteq stg(F')$. To this end, let us assume that $F' = (A', R')$ is such an AF with $\{\{b, d, f\}, \{a, c, f\}, \{a, d\}\} \subseteq stg(F')$. Using the fact that $\{b, d, f\}$ is conflict-free in $F'$ we obtain that $(d, f), (f, d) \notin R'$ and similar by using that $\{a, c, f\}$ is conflict-free in $F'$ we get that $(a, f), (f, a) \notin R'$. By assumption $\{a, d\} \in stg(F')$ and thus $\{a, d\}$ is a maximal conflict-free set of $F'$, but by the above observations the set $\{a, d, f\}$ is also conflict-free in $F'$, a contradiction. $\qquad \square$

| | $grd$ | $adm$ | $stb$ | $com$ | $prf$ | $sem$ | $stg$ |
|---|---|---|---|---|---|---|---|
| $grd$ | id | $Tr_4 \circ Tr_8$ / - | $Tr_8$ / - | $Tr_8$ / - | $Tr_8$ / ? | $Tr_8$ / ? | $Tr_8$ / ? |
| $adm$ | – | id | $Tr_6$ / - | $Tr_1$ | $Tr_4 \circ Tr_6$ / - | $Tr_6$ / - | $Tr_6$ / - |
| $stb$ | – | $Tr_4$ | id | $Tr_4$ | $Tr_4$ | $Tr_3, Tr_4$ | $Tr_3$ |
| $com$ | – | $Tr_4 \circ Tr_7$ / - | $Tr_7$ / - | id | $Tr_4 \circ Tr_7$ / - | $Tr_7$ / - | $Tr_7$ / - |
| $prf$ | – | – | – | – | id | $Tr_1$ | ? / - |
| $sem$ | – | – | – | – | – | id | ? / - |
| $stg$ | – | – | – | – | – | $Tr_2$ | id |

Table 2: Results about (weakly) faithful / exact translations.

## 6. Conclusion

In this work, we investigated intertranslations between different semantics for abstract argumentation. We focused on translations which are efficiently computable and faithful (with a few relaxations due to certain differences implicit to the semantics). An overview of our results is given in Table 2.[3] The entry in row $\sigma$ and column $\sigma'$ is to read as follows: "–" states that we have shown (Section 5) that no efficient faithful (even weakly faithful) translation for $\sigma \Rightarrow \sigma'$ exists. If the entry refers to a translation (or a concatenation of translations), we have found an efficient (weakly) exact translation for $\sigma \Rightarrow \sigma'$. An entry which is split into two parts, e.g. "$Tr_8$ / -", means that we have found an efficient (weakly) faithful translation, but there is no such exact translation. "?" indicates an open problem. We mention that all the concatenated translations are weakly faithful as they are built from a weakly exact translation $Tr_4$ (which has as only remainder set the empty set) and a faithful translation (either $Tr_6$, $Tr_7$, or $Tr_8$).

Figure 13 illustrates our intertranslatability results at one glance. Here, a solid arrow expresses that there is an efficient faithful translation while a dotted arrow depicts that there may exist such a translation, but so far we have neither found one nor have an argument against its existence. Furthermore, if for two semantics $\sigma, \sigma'$ there is no path from $\sigma$ to $\sigma'$ then it is proven (partly under typical complexity theoretical assumptions) that there is no efficient faithful translation for $\sigma \Rightarrow \sigma'$. If we consider the relations between the semantics wrt. exactness rather than just faithfulness, the overall picture changes; see Figure 14. Here, we get a more detailed picture about the relations between stable, admissible, and complete semantics. One conclusion, we can draw from these pictures is that semi-stable semantics is the most expressive one, since each of the other investigated semantics can be efficiently embedded. Moreover, we believe that our investigations complements recent results about comparisons between the different semantics proposed for argumentation frameworks.

Let us at this point also mention that, instead of considering different properties for the translations, we could also have used slightly revised semantics. The notion of remainder sets (as given in Definition 6) can partly be circumvented by, for instance, using a quasi-admissible semantics instead of admissible semantics, where the quasi-admissible extensions of an AF are all non-empty admissible extensions (in case such ones exist), or is only the empty set otherwise. Also it is obvious that the more restricted the properties for a translation are, the less such translations exist (compare Figures 13 and 14). Hence, we observe a certain trade-off between translation criteria and comparability between semantics.

---

3. One may notice that $Tr_5$ does not appear in the table. Recall that $Tr_5$ was proposed as an alternative to $Tr_2$ satisfying slightly different properties for $stg \Rightarrow sem$; see also the discussion before Theorem 14.

An alternative option to obtain translations would have been to exploit known relations between argumentation semantics and logic-programming semantics (see, e.g., Dung, 1995; Wu, Caminada, & Gabbay, 2009) and making use of known translatability results for the latter. However, we refrained from such an approach here, since it might blur the minimal requirements for the translations under consideration. In particular, from the point of meta-argumentation, translations via logic-programming semantics might introduce new arguments just for technical reasons due to the logic-programming syntax, but which have no meaning on the level of AFs.

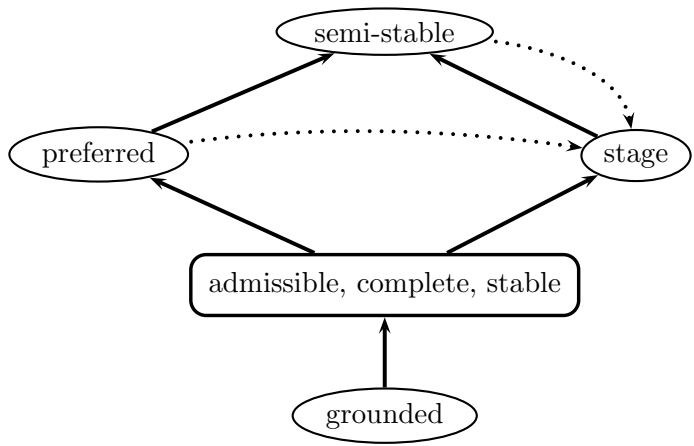

Figure 13: Intertranslatability of argumentation semantics wrt. weakly faithful translations.

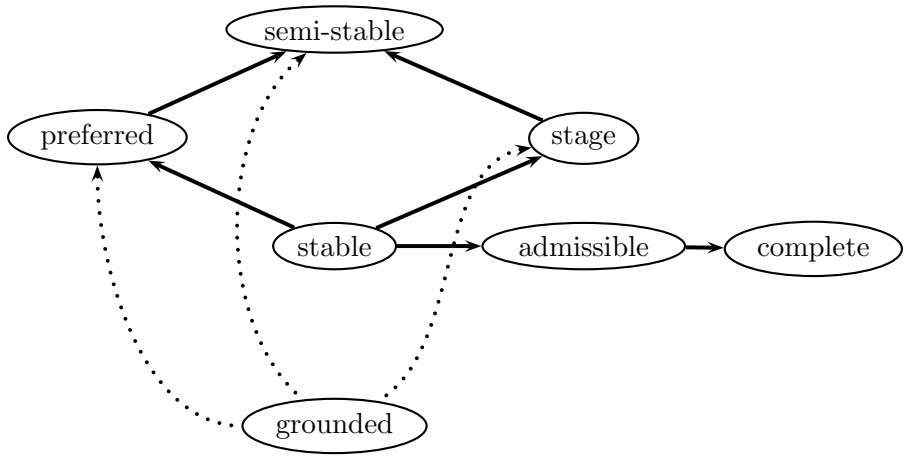

Figure 14: Intertranslatability of argumentation semantics wrt. weakly exact translations.

For future work, we identify the following tasks: First, we want to solve the few open slots in Table 2. Second, further properties for translations could be of interest. For instance, one could even strengthen the property of being exact (which is defined in terms of the extensions) to the requirement that the labelings (Caminada & Gabbay, 2009) of the source and target framework coincide. Labelings provide additional information, in particular for arguments not contained in an extension. Likewise, it would be interesting to investigate intertranslatability in the more general approach of equational semantics for argumentation frameworks (Gabbay, 2011). Further properties for translations could be given in terms of graph properties. As an example, acyclic AFs should remain acyclic after the translations, or parameters as tree-width should remain unchanged. Requirements of such a form are also termed "structural preservation" (Janhunen et al., 2006). Such properties are of interest from a computational point of view in the sense that, in case the source AF is easy to evaluate (because of its structure), this advantage should not be lost during the translation; recall here Figure 1 where we suggested to use our translations for a rapid prototyping approach to compute the extensions of a semantics via an argumentation engine based on a different semantics. Finally, we plan to extend our considerations to other important semantics like the ideal semantics (Dung et al., 2007), cf2-semantics (which is proposed among others in Baroni et al., 2005), or resolution-based semantics (Baroni, Dunne, & Giacomin, 2011), among which the resolution-based grounded semantics is of particular interest. As well studying translations between semantics for generalizations of Dung-style AFs as EAFs (Modgil, 2009) or AFRAs (Baroni, Cerutti, et al., 2011) is an interesting subject for future work.

## Acknowledgments

This work was supported by the Vienna Science and Technology Fund (WWTF) under grant ICT08-028. A preliminary version of this paper has been presented at the International Conference "*30 Years of Nonmonotonic Logic*".

The authors are grateful to Christof Spanring for suggesting the counterexample used in the proof of Theorem 15. Moreover, the authors want to thank Tomi Janhunen as well as the anonymous referees from the "*30 Years of Nonmonotonic Logic*" symposium and from JAIR for valuable comments which helped to improve the paper.